\documentclass[a4paper]{cas-sc}

\ExplSyntaxOn
\cs_gset:Npn \__first_footerline:
  { \group_begin: \small \sffamily \__short_authors: \group_end: }
\ExplSyntaxOff

\usepackage[authoryear,longnamesfirst]{natbib}
\usepackage{pgfplots}
\pgfplotsset{width=11cm,height=7cm,compat=1.3}
\usepackage{tabularx}

\usepackage{color, soul}
\sethlcolor{white}

\def\tsc#1{\csdef{#1}{\textsc{\lowercase{#1}}\xspace}}
\tsc{ASR}
\tsc{QE}
\tsc{EP}
\tsc{PMS}
\tsc{DE}
\usepackage{CJKutf8}
\usepackage[T1]{fontenc}

\begin{document}
\let\WriteBookmarks\relax
\def\floatpagepagefraction{1}
\def\textpagefraction{.001}

\shorttitle{Adapting Speech Representation Model for a New Language through Multilingual Fine-tuning and Continued Pretraining}

\shortauthors{Nowakowski et~al.}

\title [mode = title]{Adapting Multilingual Speech Representation Model for a New, Underresourced Language through Multilingual Fine-tuning and Continued Pretraining}

\author[1]{Karol Nowakowski}[auid=000,
                        bioid=1,
                        orcid=0000-0001-7435-4061]

\cormark[2]

\ead{karol@koeki-u.ac.jp}

% \ead[url]{www.cvr.cc, cvr@sayahna.org}

% \credit{Conceptualization of this study, Methodology, Software}

\affiliation[1]{organization={Tohoku University of Community Service and Science},
    % addressline={Radarweg 29}, 
    city={Sakata},
    % citysep={}, % Uncomment if no comma needed between city and postcode
    % postcode={998-0875}, 
    state={Yamagata},
    country={Japan}}

\author[2]{Michal Ptaszynski}[orcid=0000-0002-1910-9183]
\cormark[1]
\ead{michal@mail.kitami-it.ac.jp}

\affiliation[2]{organization={Kitami Institute of Technology},
    city={Kitami},
    % citysep={}, % Uncomment if no comma needed between city and postcode
    % postcode={090-8507}, 
    state={Hokkaido},
    country={Japan}}
    
\author[3]{Kyoko Murasaki}

% \credit{Data curation, Writing - Original draft preparation}

\affiliation[3]{organization={Yokohama National University (Professor Emeritus)},
    city={Yokohama},
    state={Kanagawa},
    country={Japan}}

\author%
[1]
{Jagna Nieuważny}

% \ead{}
% \ead[URL]{www.stmdocs.in}

\cortext[cor1]{Corresponding author}
\cortext[cor2]{Principal corresponding author}

\begin{abstract}
In recent years, neural models learned through self-supervised pretraining on large scale multilingual text or speech data have exhibited promising results for underresourced languages, especially when a relatively large amount of data from related language(s) is available.
While the technology has a potential for facilitating tasks carried out in language documentation projects, such as speech transcription, pretraining a multilingual model from scratch for every new language would be highly impractical.
We investigate the possibility for adapting an existing multilingual wav2vec 2.0 model for a new language, focusing on actual fieldwork data from a critically endangered tongue: Ainu.
Specifically, we
(i) examine the feasibility of leveraging data from similar languages also in fine-tuning;
(ii) verify whether the model's performance can be improved by further pretraining on target language data.
Our results show that continued pretraining is the most effective method to adapt a wav2vec 2.0 model for a new language
and leads to considerable reduction in error rates.
Furthermore, we find that if a model pretrained on a related speech variety or an unrelated language with similar phonological characteristics is available, multilingual fine-tuning using additional data from that language can have positive impact on speech recognition performance when there is very little labeled data in the target language.
\end{abstract}

% \includegraphics{figs/grabs.pdf}

% Keywords

%%%%%%%%%%%%%%%%%%%%%%%%%%%%%%%%%%%%%%%%%%%%

\maketitle

%%%%%%%%%%%%%%%%%%%%%%%%%%%%%%
\section{Introduction}
The cost of speech transcription is a major bottleneck faced in language documentation projects.
It is believed that this task could be facilitated by utilizing speech recognition technologies, but the fact that the amount of data (both annotated and unannotated) available for languages studied in such projects is typically very limited, has been a barrier in their application.

Recent studies \citep[e.g.,][]{Conneau2021UnsupervisedCR} have demonstrated the benefits of multilingual pretraining of speech representations for speech recognition in scenarios where data in the target language is scarce.
However, given the prohibitive cost of pretraining such representations from scratch, it is expected that most users will restrict themselves to fine-tuning publicly available models.
For this reason, in this paper we address the question of whether the benefits of cross-lingual transfer extend to fine-tuning, as well.
Furthermore, we study the effect of additional pretraining using the small amount of data available for the target language.
In contrast to most previous work, we focus on actual fieldwork data with all its flaws, rather than clean NLP datasets.
Specifically, our goal is to transcribe unique speech data in Sakhalin Ainu, recorded on 50-year-old audio tapes.

Our results show that continued pretraining on target language data leads to a substantial reduction in error rates.
Furthermore, we demonstrate that in a scenario where labeled data in the target language is extremely scarce, speech recognition performance can be improved by adding data from a related speech variety or an unrelated language with similar phonological traits during fine-tuning,
% but the underlying speech representation model must be first pretrained on the languages involved.
provided that the underlying speech representation model was first pretrained on that language.
Our model pretrained on the Ainu language is publicly available\footnote{\url{https://huggingface.co/karolnowakowski/wav2vec2-large-xlsr-53-pretrain-ain}}.

The remainder of this paper is organized as follows.
In the following section we discuss how speech-to-text technology and recent developments in self-supervised learning of speech representations can support language documentation.
We also introduce data on which we are focusing in this research.
In Section~\ref{sec:method}, we describe our research method.
In Section~\ref{sec:related_work}, we provide an overview of related work.
Section~\ref{sec:materials} presents the resources used in this study, including speech representation models and data used to train them.
In Section~\ref{sec:experiments}, we describe our experiments and analyze their results.
Finally, Section~\ref{sec:conclusions} contains conclusions
\hl{and ideas for future improvements.}

%%%%%%%%%%%%%%%%%%%%%%%%%%%%%%%
\section{Background}
\label{sec:background}

\subsection{Speech Transcription Technology for Language Documentation}

One of the main tasks in linguistic research concerned with endangered languages is the collection and analysis of primary linguistic data.
A typical workflow involves recording speech during fieldwork and analyzing the data afterwards. A major challenge in this process is speech transcription, which is a very time-consuming task\footnote{\hl{Depending on the annotation scheme and the level of quality required}, transcribing 1 minute of spoken language can take \hl{anywhere between several minutes and an hour} \citep{cieri-etal-2004-fisher,Gries2017LinguisticAI}.}. As a result, large amounts of data remain untranscribed in archives and collections of individual researchers. Many of those materials are stored on obsolete types of media, such as audio tapes, and in poor conditions. Before they get to be transcribed, some of them may be destroyed by accident or after the researcher who collected them retires \citep{Abney2011LanguageD}. For this reason, such materials are often referred to as ``endangered data''.
It may be possible to solve (or at least reduce) the transcription bottleneck using speech-to-text technology, thus speeding up the process of language documentation \citep{hjortnaes-etal-2020-towards,zahrer-etal-2020-towards}.
However, in order to reach high accuracy, traditional approaches require large amounts of annotated training data (on the order of thousands of hours \citep{Baevski2020wav2vec2A}), which is typically not available in a language documentation scenario.

\subsection{Cross-lingual Self-supervised Learning for Low-resource Speech Transcription}

The past few years have witnessed substantial improvements in a wide range of Natural Language Processing applications, owing to the development of efficient techniques for self-supervised learning of language representation models, such as BERT \citep{devlin-etal-2019-bert}, ELECTRA \citep{clark2020electra} (for text-based models), wav2vec 2.0 \citep{Baevski2020wav2vec2A} and HuBERT \citep{Hsu2021HuBERTSS} (for speech representations).
They have been shown to produce competitive results compared to traditional, fully supervised methods, while training on much fewer human-annotated samples (after having been pretrained on large amount of unlabeled data).

A major obstacle for applying self-supervised learning in a language documentation setting is the fact
that for the majority of the world's languages, even raw text or speech data is not available in large quantities.
To alleviate this problem, recent studies explore cross-lingual learning techniques \citep{conneau-etal-2020-unsupervised,Singh2019XLDACD}.
It has been demonstrated that learning a single model from unlabeled data in multiple languages can have positive impact on the quality of representations computed for each individual language.
As an example,
\cite{Conneau2021UnsupervisedCR} pretrained their speech representation model on 56k hours of unlabeled data in 53 languages and found it to perform far better than monolingual models, particularly for languages with little data available.
However, the cost of compute required to train such models and the energy consumed in doing so is extremely high\footnote{For instance, \cite{Conneau2021UnsupervisedCR}'s multilingual model was trained using 64 GPUs, while \cite{conneau-etal-2020-unsupervised} trained their XLM-R model with 500 GPUs.}.

\subsection{Sakhalin Ainu Speech Data}
\label{sec:tapes}

The aim of this research is to develop a system for automatic transcription of Ainu, a critically endangered language\footnote{From the second half of the 20\textsuperscript{th} century Ainu has not been used as a language of everyday communication \citep{bugaeva2012}, thus many specialists consider it extinct. There are, however, efforts to revitalize the language and a growing number of people are learning it.} native to northern Japan, Sakhalin and Kuril Islands.
In particular, we are focusing on the task of transcribing unpublished materials from several dialects of the Ainu language spoken in Sakhalin, recorded in the 1960s and 1970s by Professor Ky\={o}ko Murasaki, with some of the last speakers of those dialects: Haru Fujiyama (Rayciska dialect), \={O}ta Yuku (Maoka dialect), Chikama Kimura (Shirahama dialect) and others.
The materials in question were originally recorded on over 30 magnetic tapes of an old type (so called ``reel-to-reel'' tapes or ``open-reel'' tapes), and were a good example of ``endangered data''.
Sound quality is rather poor, with high levels of noise \hl{(both equipment noise such as hiss and hum, as well as occasional ambient noise)} and a considerable amount of distortions.
\hl{The bulk of the recorded materials consists of spontaneous monologues (mainly reciting folktales) by a single speaker, but there are also conversations between the informant and the interviewer or involving multiple informants, and occasionally multiple participants speak simultaneously.
The tapes were recorded mainly at the home of one of the informants using a portable recorder.}

The total duration of the recordings exceeds 20 hours, which -- to the best of our knowledge -- makes it larger than any collection of Sakhalin Ainu texts published so far\footnote{In addition to Sakhalin Ainu, the tapes in question also contain 2 hours of speech recordings in Hokkaido Ainu.}.
A subset of the recordings has been transcribed, translated to Japanese and published:
\cite{Murasaki1976} released a collection of eleven folktales by Haru Fujiyama and five short conversations between two native speakers (Fujiyama and \={O}ta).
\cite{Murasaki2010} produced a compilation of three different versions of a single folktale, ``Wenenekaype'',
recited by Haru Fujiyama.
Lastly, two volumes by \cite{Murasaki2013_sentences1} and \cite{Murasaki2016_sentences2} contain a total of 297 sentences.

Sakhalin Ainu is one of the three major dialect groups recognized within the Ainu language (the other two being Hokkaido Ainu and Kuril Ainu).
There are significant differences between the dialects of Sakhalin and Hokkaido and they are mutually unintelligible \citep{refsing1986,Murasaki2009} (Table~\ref{tab:ainu_sentences} includes examples of semantically equivalent sentences in both speech varieties).
\cite{Vovin2016} describes Ainu as a portmanteau language family with two primary branches: Hokkaido-Kuril and Sakhalin.
In the case of Hokkaido Ainu, a sizeable body of audio and written materials have been recorded and published, and recently a growing number of them are digitized and released online.
Data in Sakhalin Ainu, however, is far more scarce\footnote{For Kuril Ainu, there is almost no data available, apart from several lexicons and word lists.}.

\begin{table}[]
\caption{\hl{Examples of equivalent sentences in Sakhalin Ainu (Rayciska dialect) and Hokkaido Ainu (Horobetsu dialect) recorded in} \cite{hattori1964}. \hl{For comparison, we also include Japanese translations (romanized according to the Hepburn transliteration system).}}
\label{tab:ainu_sentences}
\centering
\begin{tabularx}{\textwidth}{>{\raggedright}X>{\raggedright}XXX}
\toprule
Sakhalin Ainu & Hokkaido Ainu & Japanese & English \\
\midrule
ku'ani 'enahkari nee 'aynuka hennehka 'oman. & 'enmosma nen ka senne 'oman. & watashi no hoka ni wa dare mo ikanai. & There's no one going but me. \\
\midrule
'uneeno 'an 'itahpateh kisci. & sine 'itak 'ukoraci 'an 'itak patek ye kor an. & onaji koto bakari itte iru. & He always says the same thing. \\
\midrule
tah 'aynu'itakani temana ayyeepe? & tanpe 'anak 'aynu'itak 'ari nekona 'aye ya. & kore wa ainugo de nan to iimasu ka? & What is this called in Ainu? \\
\bottomrule
\end{tabularx}
\end{table}

\hl{
Until now, none of the numerous hypotheses about genetic relationships between Ainu and other languages or language families has gained wider acceptance, and thus it is usually classified as a language isolate.
In terms of linguistic typology, Ainu is an agglutinating language with SOV (subject-object-verb) word order and elements of polysynthesis, such as noun incorporation and concentration of various morphemes in the verbal complex} \citep{shibatani1990}.
\hl{
Phonemic inventory of Ainu consists of five vowel phonemes: /i, e, a, o, u/, and~twelve consonant phonemes: /p, t, k, c, s, h, r, m, n, y, w, '/ (/'/ denotes a glottal stop).
Syllables in Sakhalin Ainu conform to one of the following patterns: CV, CVV (VV represents a long vowel) or CVC.
Most written texts in Ainu are transcribed using Latin alphabet and/or an extended version of the Japanese \textit{katakana} syllabary (textual data in Ainu used in this research is written in Latin script).
The majority of contemporary experts
follow the phonemic orthographic rules devised by
} \cite{hattori1964} \hl{or a slightly modified version proposed by} \cite{akor_itak1994}.
\hl{However, certain aspects of the writing system, such as word segmentation, have not been standardized (for more details, please refer to} \cite{Nowakowski2019MiNgMatchA}).

\section{Research Method}
\label{sec:method}

Given the encouraging results achieved by multilingual speech representation models for underresourced languages, and the high cost of pretraining such a model from scratch, we explore two methods for adapting an existing model for a new language with a limited amount of available speech data.

\smallskip
\noindent\textbf{Multilingual fine-tuning} In addition to multilingual self-supervised pretraining, \cite{Conneau2021UnsupervisedCR} conducted an experiment with fine-tuning using labeled data from 10 different languages simultaneously, and found the resulting model to perform competitively to models fine-tuned on each language individually.
On the other hand, they did not analyze the correlation between language similarity and effectiveness of such multilingual fine-tuning\footnote{They did perform such analysis for pretraining, and found that low-resource language performance benefits more from additional data in similar languages.}.
In this study, we examine whether the benefits of cross-lingual transfer between closely related languages/dialects or unrelated languages sharing some phonological characteristics, apply to fine-tuning as well.
To that end, we carry out fine-tuning experiments using data in the target language (Sakhalin Ainu) in combination with relatively large amounts of data from three different speech varieties: Hokkaido Ainu, Japanese and English.
As mentioned in Section~\ref{sec:tapes}, Sakhalin Ainu and Hokkaido Ainu can be viewed as distant varieties of the same language or as closely related languages.
As for Japanese, the theory of a genetic relationship between it and the Ainu language(s) is rejected by most experts \citep{refsing1986,shibatani1990}.
That being said, and despite substantial differences in such aspects as consonants allowed in syllable coda and accent, the phonological system of Ainu has arguably more in common with Japanese\footnote{Some of the similarities are presumably a result of contact-induced change \citep{bugaeva2012}.}\textsuperscript{,}\footnote{See \cite{nowakowski2020rall} for an example of using Japanese speech models to recognize and generate speech in Ainu. \cite{Matsuura2020SpeechCO} trained an end-to-end ASR model for Hokkaido Ainu using additional Japanese and English data and found the former to be more helpful.} than, e.g., English.
This intuition is also corroborated by the analysis of language vectors computed using \texttt{lang2vec} \citep{littell2017uriel}\footnote{\url{https://github.com/antonisa/lang2vec}}: upon calculating the distances between the Ainu language and all other languages in the database (specifically, we compared \texttt{phonological} and \texttt{inventory} features and took the mean distance), we found that Japanese is in 43\textsuperscript{rd} position in terms of proximity to Ainu, out of 8070 languages.

\smallskip
\noindent\textbf{Continued pretraining} Secondly, we investigate if it is possible to improve the performance of a strong multilingual model on a language not seen during initial pretraining, by performing additional pretraining on small amount of target language data.

\section{Related Work}
\label{sec:related_work}

\cite{Schneider2019wav2vecUP} introduced wav2vec, a technique for self-supervised learning of speech representations from raw audio data using convolutional neural networks trained to distinguish true audio samples from distractors.
Their approach outperformed the previous state-of-the-art on the WSJ speech recognition benchmark while using two orders of magnitude less labeled data.
\cite{vq-wav2vec} extended their work by adding a quantization module for computing discrete representations of audio segments and feeding the discretized sequence to a Transformer (BERT) model.
Further improvements were introduced by \cite{Baevski2020wav2vec2A} who proposed wav2vec 2.0, an end-to-end framework for jointly learning discretized speech units and contextualized speech representations, and fine-tuned the resulting model for speech transcription instead of feeding the pretrained features to a separate downstream model.
\cite{Conneau2021UnsupervisedCR} pretrained a single wav2vec 2.0 model (dubbed XLSR-53) using 56k hours of speech data in 53 languages and obtained a higher accuracy in speech recognition than monolingual models or previous methods.
\cite{Hsu2021RobustW2} combined speech data from different domains and investigated the impact of domain mismatches in self-supervised learning for ASR.
\cite{Xu2021SimpleAE} performed zero-shot transcription of unseen languages by fine-tuning the XLSR-53 model and mapping phonemes of the training languages to the target language using articulatory features.
\cite{babu2021xlsr} used wav2vec 2.0 and 436k hours of unlabeled data in 128 languages to train large-scale (up to 2 billion parameters) models, which after fine-tuning achieved state-of-the-art performance in speech recognition, speech translation and language identification.
\cite{Sriram2022Wav2VecAugIS} \hl{obtained improved ASR performance by applying data augmentation techniques -- such as pitch shift and adding random noise to the input signal -- to the pretraining data, and introducing several modifications to the wav2vec 2.0 architecture.}
\cite{Sanabria2022MeasuringTI} \hl{used models pretrained on modified natural speech or synthetic data to measure the impact of individual domain factors (vocabulary, word order, phonetic features, etc.).
They found that low-level domain factors, such as phonotactics and prosody, play a more important role than syntactic or lexical variation, and that speaker diversity in the pretraining data is crucial.
Furthermore, they demonstrated that using a large amount of synthesized data can lead to better performance than with the small amount of real data used to train the synthesizer.}
\cite{Wu2022} and \cite{Vyas2022OndemandCR} \hl{proposed modifications to the wav2vec 2.0 architecture aimed at reducing the computational cost of pretraining and inference.}

Previous studies found continued self-supervised training of textual language representation models to be an effective method for adapting them for a new domain \citep{howard-ruder-2018-universal,Sun2019,gururangan-etal-2020-dont} or expanding their coverage to languages unseen in initial pretraining \citep{pfeiffer-etal-2020-mad,Tang2020MultilingualTW,ebrahimi-kann-2021-adapt}.
Further pretraining of a speech representation model on new languages was investigated
by \cite{Kessler2021Continualwav2vec2AA}.
However, they only conducted experiments with a monolingual (English) model and focused on a high-resource setting, where the available speech data in the newly added language is ample (800 hours or more).
\cite{Khurana2022} \hl{used self-training to adapt monolingual English wav2vec 2.0 models for several other languages in a simulated low-resource scenario.} 
Multilingual fine-tuning was studied in the context of text-based machine translation by \cite{Tang2020MultilingualTW} and resulted in improved performance, especially on low-resource languages.

\section{Materials}
\label{sec:materials}

\subsection{Wav2vec 2.0}

In all speech transcription experiments described in this paper, a publicly available pretrained wav2vec 2.0 model was employed.
Specifically, we used the XLSR-53 -- a model trained on 56k hours of data in 53 languages -- compiled and released\footnote{\url{https://github.com/pytorch/fairseq/tree/main/examples/wav2vec}} by \cite{Conneau2021UnsupervisedCR}.
Furthermore, we performed an additional pretraining of the XLSR-53 model on Ainu language data described in the next section.
Both pretraining and fine-tuning of the model were conducted using the \texttt{fairseq} library.

%%%%%%%%%%%%%%%%%%%%%%%%%%%%%%%%%
\subsection{Pretraining Data}

In the experiments with continued pretraining of the XLSR-53, we used a total of 234 hours of speech data in Sakhalin Ainu and Hokkaido Ainu\footnote{While it might be informative to also pretrain a model on Sakhalin data only, we decided not to do so, due to the high cost of pretraining.}. Specifically, we pretrained our model on data obtained from tapes described in Section~\ref{sec:tapes} and from publicly available data collections listed in Table~\ref{tab:pretr_data}.

\begin{table}[width=1.0\linewidth,cols=3,pos=htb]
\caption{Statistics of the data used for continued pretraining. In the case of the Sakhalin Ainu data obtained from tapes, we included duplicate recordings, hence two numbers are reported (the number in brackets corresponds to the duration of unique recordings).}\label{tab:pretr_data}
\begin{tabularx}{\textwidth}{Xll}
\toprule
Data & (Main) language/dialect & Total duration (h)\\
\midrule
Sakhalin Ainu tapes & Sakhalin Ainu & 35.9 (21.5)  \\[5pt]
Tuytah \citep{Murasaki2001_Tuytah} & Sakhalin Ainu & 8.9 \\[5pt]
``Wenenekaype'' \citep{Murasaki2010} & Sakhalin Ainu & 1.9 \\[5pt]
Ainu Language Archive \citep{ainu_archive} & Hokkaido Ainu & 103.1 \\[15pt]
Dictionary of Mukawa Ainu \citep{mukawa} & Hokkaido Ainu & 26.5 \\[15pt]
Ainu Language \& Ainu Oral Literature \citep{nibutani_literature} & Hokkaido Ainu & 19.2 \\[15pt]
Ainu language audio materials in the Waseda University Repository (\url{https://waseda.repo.nii.ac.jp/}) & Hokkaido Ainu & 14.1 \\[15pt]
ILCAA's Project for the Publication of Ainu Language Materials \citep{AA_ken_Ainugo_shiryo} & Hokkaido Ainu & 12.1 \\[35pt]
Glossed Audio Corpus of Ainu Folklore \citep{glossed_corpus_of_ainu} & Hokkaido Ainu & 6.2 \\[5pt]
Shigeru Kayano's Ainu dictionary \citep{kayano1996} & Hokkaido Ainu & 3.0 \\[5pt]
A Topical Dictionary of Conversational Ainu \citep{topicaldictionaryofainu} & Hokkaido Ainu & 2.3 \\[15pt]
\textit{Ny\={u} ekusupuresu Ainugo} \citep{nakagawa2013} & Hokkaido Ainu & 1.0 \\[5pt]
\bottomrule
\end{tabularx}
\end{table}

\begin{table}[width=1.0\linewidth,cols=3,pos=thp]
\caption{Statistics of the speech data and transcriptions used for fine-tuning. \hl{Japanese is written without spaces and we did not perform tokenization, hence we don't report token counts and vocabulary sizes for the Japanese data.}}\label{tab:ft_data}
\begin{tabularx}{\textwidth}{Xllll}
\toprule
Data & (Main) & Total & \hl{Token} & \hl{Vocab.}\\
& language/dialect &  duration (h) & \hl{count} & \hl{size}\\
\midrule
Tuytah \citep{Murasaki2001_Tuytah} & Sakhalin Ainu & 8.9 & 52,172 & 4,415 \\[5pt]
``Wenenekaype'' (\texttt{Fu12-690401}) \citep{Murasaki2010} & Sakhalin Ainu & 0.8 & 5,817 & 1,165 \\[15pt]
Ainu Language Archive \citep{ainu_archive} & Hokkaido Ainu & 62.2 & 396,755 & 10,618 \\[15pt]
A Topical Dictionary of Conversational Ainu \citep{topicaldictionaryofainu} & Hokkaido Ainu & 2.3 & 13,007 & 2,260 \\[15pt]
Common Voice (Japanese) \citep{commonvoice:2020} & Japanese & 40.6 & N/A & N/A \\[5pt]
JSUT \citep{Sonobe2017JSUTCF} & Japanese & 10.3 & N/A & N/A  \\[5pt]
LibriSpeech \citep{Librispeech} & English & 100.6 & 990,101 & 33,798 \\
\bottomrule
\end{tabularx}
\end{table}

The recordings from open-reel tapes were transferred to a digital format (WAV) using an audio recorder connected to a tape deck.
The tapes are double-sided and we found that many of the recordings were audible in data obtained from both sides of the same tape.
Upon inspection it turned out that in some instances, data retrieved from the reverse side (or certain parts of it) is superior to the corresponding data on the front side in terms of quality of the audio signal.
For this reason, we included the duplicate recordings in the pretraining data, hence two numbers are reported in Table~\ref{tab:pretr_data} (the number in brackets corresponds to the duration of unique recordings).

All files were converted to a single channel WAV sampled at 16 kHz.
Files longer than 15 seconds were automatically split on silence intervals (using \texttt{pydub}) into separate clips 2 to 15 seconds in length.
Files shorter than 1 second were excluded from pretraining.

\subsection{Fine-tuning Data}

Data used for fine-tuning is listed in Table~\ref{tab:ft_data}.
In monolingual fine-tuning, we used Sakhalin Ainu data from two sources: one story from \cite{Murasaki2010} (namely, \texttt{Fu12-690401}; the remaining two recordings were used for validation and testing) and data from \cite{Murasaki2001_Tuytah}\footnote{Also available online at: \url{http://www.aa.tufs.ac.jp/~mmine/kiki_gen/murasaki/asai01.html}}.
In multilingual fine-tuning, we added data from Hokkaido Ainu (64.5h), Japanese (validated subset of the Japanese data in the Common Voice Corpus 8.0 \citep{commonvoice:2020}\footnote{\url{https://commonvoice.mozilla.org/ja/datasets}}, version \texttt{ja\_43h\_2022-01-19}, and the JSUT corpus \citep{Sonobe2017JSUTCF}\footnote{\url{https://sites.google.com/site/shinnosuketakamichi/publication/jsut}}
; 50.9h in total) and English (the 100h ``clean'' subset of LibriSpeech).

Speech data obtained from \cite{Murasaki2010} was automatically split on silence intervals (using \texttt{pydub}) into separate clips 2 to 15 seconds in length.
Transcriptions from the book were digitized and aligned with the audio clips.

All audio files were converted to a single channel WAV sampled at 16 kHz.
Punctuation marks and metadata were removed from all transcriptions.
All alphabetic characters in the transcriptions of Sakhalin Ainu and Hokkaido Ainu texts were converted to lower case.
In order to prevent a large increase in the output vocabulary size and reduce data sparsity, transcriptions in Japanese were transliterated (using \texttt{pykakasi}) to the \textit{katakana} syllabary. This also applies to data in the Ainu language which contains many words and utterances in Japanese (code-switching, comments about the text, questions from an interviewer, etc.).
LibriSpeech transcriptions were used in their original form, i.e., in all upper case letters.

An excerpt from the transcriptions for ``Wenenekaype'' is shown in Table~\ref{tab:ft_data_sample}.

\begin{table}[htpb]
\centering
\caption{Excerpt from the transcriptions for “Wenenekaype”, used in fine-tuning of our speech recognition models.}
\begin{tabularx}{\textwidth}{XXX}
\toprule
Original transcription \citep{Murasaki2010} & Preprocessed for fine-tuning & English translation \\
\midrule
sine, sine.. 'oyanruru kotan 'ohta sine, 'oyanruru kotan 'an manu. 'an manuyke reekoh wenporo kotan 'an manu. 'ani ike, hemanta ka, hemanta 'oyasi hee, 'an manuyke neyan wenporo kotan 'oma 'aynu ka 'emuyke ruhpa 'ike tuy wa (tuy)pa wa 'isam. 'isam mayne tani 'ampene 'oha kotan nee manu. & sine sine 'oyanruru kotan 'ohta sine 'oyanruru kotan 'an manu 'an manuyke reekoh wenporo kotan 'an manu 'ani ike hemata ka hemata'oyasi hee 'an manuyke neyan wenporo kotan 'oma 'aynu ka 'emuyke ruhpa nike tuy wa tuypa wa 'isam 'isam mayne tani 'ampene 'oha kotan nee manu & There was a big village. A really big village. But then, there was some kind of monster that ate away all the people in that big village, and there was no one left. They're all gone, and now it's just an empty village. \\
\bottomrule
\end{tabularx}
\label{tab:ft_data_sample}
\end{table}

\section{Experiments}
\label{sec:experiments}

%%%%%%%%%%%%%%%%%%%%%%
\subsection{Continued Pretraining}
\label{sec:cont_pretr}

A random subset of 1\% of the data was used for validation.
Pretraining was performed using four Nvidia GTX 1080Ti GPUs.
We continued pretraining for 100k updates, which took a total of 5 weeks.
A small learning rate (1e-4, compared to 1e-3 used by \cite{Conneau2021UnsupervisedCR} in initial pretraining) was set to prevent catastrophic forgetting \citep{Sun2019}.
We used a batch size of 150k samples per GPU and applied gradient accumulation to simulate 512 GPUs, reaching an effective batch size of 80 minutes.
Other hyperparameters were set according to the configuration for the \textsc{\lowercase{LARGE}} model reported by \cite{Baevski2020wav2vec2A}.

\subsection{Fine-tuning}

We fine-tuned the pretrained models for speech transcription with a CTC loss \citep{Graves_CTC}.
The best checkpoint for each experiment run was selected according to Word Error Rate on the validation set (for this purpose, we used the shortest, 10-minute recording from \cite{Murasaki2010}, namely, \texttt{Fu13-700326}).
The output of the fine-tuned models was decoded with a Viterbi decoder and a 4-gram language model trained on the Sakhalin Ainu part of the data used in fine-tuning of the corresponding model (see Table~\ref{tab:lm_stats}).
Language models were computed using the KenLM toolkit\footnote{\url{https://kheafield.com/code/kenlm/}}.
The performance in speech transcription was evaluated on a 37-minute subset of the data from \cite{Murasaki2010} (namely, \texttt{Fu11-690328}).
Before the evaluation, transcriptions generated by the system were preprocessed by converting all alphabetic characters to lower case.
We report Character Error Rate (CER) and Word Error Rate (WER).

\begin{table}[width=1.0\linewidth,cols=5,pos=tp]
\caption{\hl{Statistics of the KenLM language models used for decoding, including perplexity and out-of-vocabulary token rates on the evaluation data.}}\label{tab:lm_stats}
\begin{tabular*}{\tblwidth}{@{} LLLLL@{} }
\toprule
Training & Vocabulary & Perplexity & Perplexity & Out-of-vocabulary \\
data & size & (including OOVs) & (excluding OOVs) & token rate \\
\midrule
``Wenenekaype'' & 1,168 & 152.1 & 92.9 & 563/4911 (11.5\%) \\
``Wenenekaype'' + Tuytah & 5,095 & 290.7 & 181.9 & 403/4911 (8.2\%) \\
\bottomrule
\end{tabular*}
\end{table}

All fine-tuning experiments were conducted using a single Nvidia RTX 3090 GPU.
The baseline models were fine-tuned for 15k updates on the few labeled samples from the target domain (i.e., ``Wenenekaype'') only, with a batch size of 2.56M samples and the learning rate set to 3e-4.
Other hyperparameters were set in accordance with the configuration for the \textsc{\lowercase{LARGE}} model reported by \cite{Baevski2020wav2vec2A}.
On our system, fine-tuning with these settings took less than 1.5h to complete. 
After adding more data from Sakhalin Ainu, we fine-tuned for 20k steps with a learning rate of 1e-4.
While the batch size was reduced to 800k samples, we applied gradient accumulation to simulate 16 GPUs, which resulted in an effective batch size of 12.8M, and training time of around 10 hours.
In this setting, the ``Wenenekaype'' data was oversampled by a factor of 10 (we found that without oversampling, performance on the validation set was much worse).
For models fine-tuned on a combination of data from two speech varieties, we set the learning rate to 3e-5 and executed 80k updates.
When fine-tuning on Sakhalin Ainu and Hokkaido Ainu data together, the ``Wenenekaype'' data was oversampled by a factor of 200 and the Tuytah data was oversampled by a factor of 10.
In experiments not using the latter data, ``Wenenekaype'' was oversampled by a factor of 100.
In this case, a single fine-tuning run took roughly 2 days.
After adding data from a third language (i.e., Japanese), we fine-tuned for 120k steps and almost 3 days.
Finally, when fine-tuning on data from all four languages, we executed 160k updates, which took 4 days.
In both cases, the ``Wenenekaype'' data was oversampled to comprise roughly half of the training data.

%%%%%%%%%%%%%%%%%%%%%%
\subsection{Results and Analysis}
\label{sec:results}

\begin{figure}[htp]
    \centering

    \begin{tikzpicture}

      \pgfplotsset{
          scale only axis,
      }
    
      \begin{axis}[
        axis y line*=left,
        xlabel={Pretraining steps / 1000},
        ylabel={CER},
        ymin=10,
        xtick={0,10,20,30,40,50,60,70,80,90,100},
        grid=both,
        ymajorgrids=true,
        grid style=densely dotted,
      ]

        \addplot[mark=square, color=orange]
          coordinates{
            (0,13.9)(10,12.1)(20,11.4)(30,11.2)(40,10.5)(50,10.8)(60,10.3)(70,10.8)(80,10.7)(90,10.5)(100,10.6)
          }; \label{plot_cer}

        \addplot[mark=o, color=red]
          coordinates{
            (0,18.0)(10,16.0)(20,15.2)(30,15.7)(40,14.5)(50,15.1)(60,14.2)(70,14.3)(80,14.7)(90,14.7)(100,14.6)
          }; \label{plot_cer_lm}
    
        \end{axis}
    
        \begin{axis}[
          axis y line*=right,
          axis x line=none,
          ylabel={WER},
          ymin=30,
          ymajorgrids=true,
          grid style=dashed
        ]
        
        \addlegendimage{/pgfplots/refstyle=plot_cer}\addlegendentry{CER (Viterbi)}
        \addlegendimage{/pgfplots/refstyle=plot_cer_lm}\addlegendentry{CER (KenLM)}

        \addplot[mark=asterisk, color=cyan]
          coordinates{
            (0,42.3)(10,38.0)(20,36.0)(30,35.1)(40,33.2)(50,34.2)(60,32.7)(70,34.3)(80,33.3)(90,32.5)(100,33.5)
          }; \addlegendentry{WER (Viterbi)}
          
        \addplot[mark=triangle, color=blue]
          coordinates{
            (0,36.5)(10,34.2)(20,33.1)(30,33.2)(40,32.3)(50,32.9)(60,32.5)(70,32.9)(80,32.9)(90,32.7)(100,33.0)
          }; \addlegendentry{WER (KenLM)}

      \end{axis}

    \end{tikzpicture}

    \caption{Effect of further pretraining using target language data. The models were fine-tuned using ``Wenenekaype'' data only.}
    \label{fig:effect_of_pretr}
\end{figure}
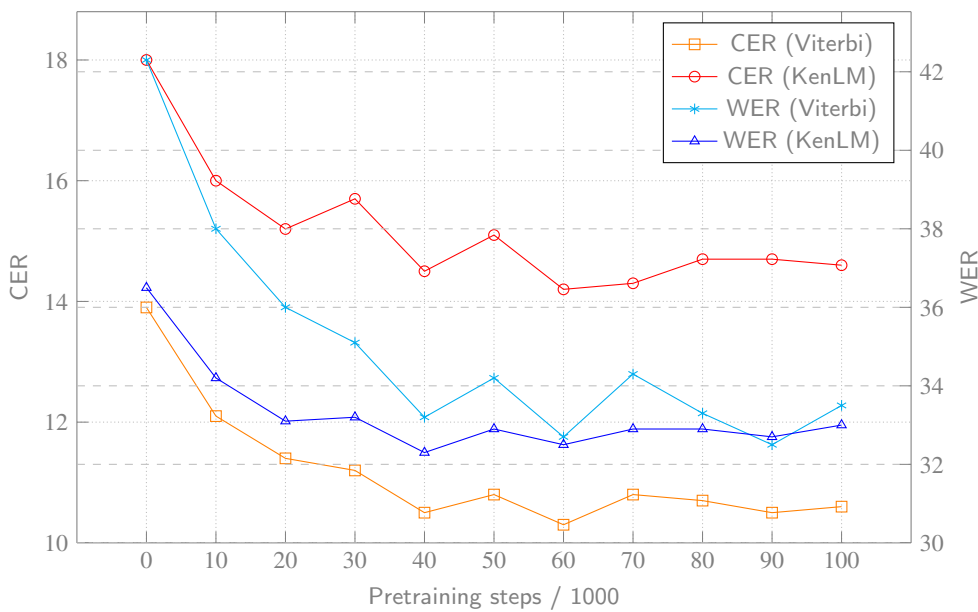

In Table~\ref{tab:ft_rslts} we compare error rates yielded by models fine-tuned on monolingual or multilingual data, and with or without additional pretraining on Ainu data, in speech recognition on the test set.
Figure~\ref{fig:effect_of_pretr} provides an analysis of the impact of additional pretraining using target language data.

\begin{table}[width=.9\linewidth,cols=6,pos=tp]
\caption{Comparison of models fine-tuned on monolingual or multilingual data, with or without additional pretraining on target language data, in speech transcription on ``Wenenekaype'' test set. We report Character Error Rates and Word Error Rates.}\label{tab:ft_rslts}
\begin{tabular*}{\tblwidth}{@{} LLLLLL@{} }
\toprule
& Additional & \multicolumn{2}{c}{Viterbi} & \multicolumn{2}{c}{KenLM} \\
Fine-tuning data & pretraining & \multicolumn{2}{c}{decoder} & \multicolumn{2}{c}{decoder} \\
& steps & CER & WER & CER & WER\\
\midrule
``Wenenekaype'' & 0 & 13.9 & 42.3 & 18.0 & 36.5 \\
Sakhalin Ainu (``Wenenekaype'' + Tuytah) & 0 & 11.8 & 37.3 & 14.1 & 31.8 \\
Sakhalin Ainu + Hokkaido Ainu & 0 & 15.9 & 50.6 & 20.2 & 38.6 \\
``Wenenekaype'' + Hokkaido Ainu & 0 & 16.4 & 51.1 & 21.8 & 41.2 \\
``Wenenekaype'' + Japanese & 0 & 15.7 & 41.8 & 19.0 & 38.0 \\
``Wenenekaype'' + English & 0 & 13.9 & 43.0 & 17.5 & 38.0 \\
``Wenenekaype'' + Hokk. Ainu + Jap. & 0 & 16.0 & 52.0 & 22.1 & 41.1 \\
\hl{``Wenenekaype'' + Hokk. Ainu + Jap. + Eng.} & 0 & 15.2 & 47.4 & 20.5 & 40.4 \\
``Wenenekaype'' & 100k & 10.6 & 33.5 & 14.6 & 33.0 \\
Sakhalin Ainu (``Wenenekaype'' + Tuytah) & 100k & \textbf{9.7} & \textbf{29.8} & 13.4 & 30.2 \\
Sakhalin Ainu + Hokkaido Ainu & 100k & 10.1 & 31.1 & 12.6 & 30.8 \\
``Wenenekaype'' + Hokkaido Ainu & 100k & 10.2 & 30.5 & 15.1 & 32.1 \\
``Wenenekaype'' + Japanese & 100k & 10.8 & 32.2 & 15.2 & 33.0 \\
``Wenenekaype'' + English & 100k & 12.0 & 40.1 & 16.8 & 34.2 \\
``Wenenekaype'' + Hokk. Ainu + Jap. & 100k & 10.9 & 33.7 & 16.2 & 33.4 \\
\hl{``Wenenekaype'' + Hokk. Ainu + Jap. + Eng.} & 100k & 10.9 & 33.9 & 16.8 & 34.3 \\
\bottomrule
\end{tabular*}
\end{table}

Our results show that continued pretraining is clearly the most effective way to adapt a speech representation model for a new language.
With just 10k updates (half a week on our system), we were able to obtain a reduction in CER by nearly 13\% (when decoding the model's output with a Viterbi decoder).
After two weeks and 40k updates, we reached an improvement by 24.5\%.
The best relative performance on our test data in terms of CER was obtained after 60k updates.
Concerning WER, the lowest values were measured after 60k and 90k updates (32.7 and 32.5, respectively) when decoding with a Viterbi decoder, and after 40k steps (32.3) when using a language model.

Fine-tuning on all the available Sakhalin Ainu data and decoding without a language model yielded the best overall results.
With the model before conducting further pretraining, multilingual fine-tuning was not helpful, regardless of which language combination was used and how much target language data was available (the addition of Japanese data did result in slightly lower WER, but at the cost of an increase in CER).
Quite surprisingly, adding Hokkaido Ainu data not only had negative impact on the model's performance, but also resulted in significantly higher error rates than in experiments using data in Japanese and English.
There are two possible reasons for this behavior: (i) not having learnt a representation for either of the Ainu languages, the model was unable to take advantage of their similarities, and (ii) Japanese and English are distant enough from the target language for the model to be able to easily discern between them, thus making relatively few errors due to confusion between languages.

%%%%%%%%%%%%%%%%%%%%%%
After additional pretraining, the results of multilingual fine-tuning were completely different:
adding data from another variety of Ainu produced the best performing model, followed by fine-tuning on unrelated languages sharing some phonological characteristics (i.e., Ainu and Japanese), whereas combining Ainu with English was clearly harmful.
While additional data from Hokkaido Ainu did not lead to an improvement compared to the model trained on 10 hours of Sakhalin data, the results did improve in the scenario with less than 1 hour of target language data available.
In this case, a drop in WER was also observed after including Japanese data.

In experiments using labeled data from two or three additional languages, we observed worse results, even with languages which were helpful when used individually (namely, Hokkaido Ainu and Japanese).
We think that this behavior might be associated with lower capacity in the fine-tuned model for each of the languages, but it requires further analysis.

Analysis of the transcriptions generated when using a Viterbi decoder
revealed that some of the errors were caused by an incorrect choice of the writing system (i.e., the use of \textit{katakana} characters to represent sounds of the Ainu language or, conversely, transcribing parts spoken in Japanese using Latin alphabet letters; an example is shown in Table~\ref{tab:ex_errors_jp}).
In fact, linguists often transcribe Japanese code-switched words found in Ainu language texts using a romanization system, rather than the Japanese script.
For this reason, we decided to examine how the results would change if we relax our problem by converting all Japanese characters in both the system's output and ground truth data to Latin alphabet.
Specifically, we used \texttt{pykakasi} to romanize Japanese according to the Hepburn transliteration system.
Results are presented in Table~\ref{tab:ft_rslts_all_latin}.
While the modification resulted in a slightly lower CER in almost all configurations, the biggest improvements were observed for models fine-tuned using Japanese data.
Under this evaluation scheme, it can be concluded that fine-tuning jointly on ``Wenenekaype'' data and Japanese speech data is clearly beneficial.

\begin{CJK}{UTF8}{ipxm}
\begin{table}[]
\caption{Excerpt from the transcriptions generated by a fine-tuned model, showing errors caused by an incorrect choice of the writing system.}
\label{tab:ex_errors_jp}
\centering
\begin{tabularx}{\textwidth}{rX}
\toprule
Model output:  & tanna 'an 'opompaki nah ramupe オota'asi nee manu 'タ 'asi ソ'oヤw nah 'anramu 'ampe アノ ジンkoy nee manu \\
Transliteration: & [tanna 'an 'opompaki nah ramupe oota'asi nee manu 'ta 'asi so'oyaw nah 'anramu 'ampe ano jinkoy nee manu] \\
\midrule
Ground truth:  & tani neya アノ 'opompaki nah ramupe 'ota'asi nee manu 'ota'asi suy 'oyaw nah 'anramu'ampe アノ cinkoy nee manu \\
Transliteration: & [tani neya ano 'opompaki nah ramupe 'ota'asi nee manu 'ota'asi suy 'oyaw nah 'anramu'ampe ano cinkoy nee manu] \\
\bottomrule
\end{tabularx}
\end{table}
\end{CJK}

The above results are consistent with our hypothesis that labeled data from similar languages can be leveraged in fine-tuning to obtain better performance on the target language.
On the other hand, they indicate that any improvement can only take place if the following two conditions are met: (i) the model was first pretrained on the languages involved in fine-tuning, or at least on one of them (namely, the transfer language whose data one intends to use in addition to the target language data), and (ii) the amount of labeled data in the target language is extremely low.

\hl{Some of the errors made by the system can be attributed to inconsistencies in the annotations used for fine-tuning and evaluation, namely, differences between the transcriptions of multiple instances of the same lexical item (which in turn, are a result of the absence of a standardized orthography for the Ainu language).
For instance, the part of ``Wenenekaype'' used for evaluation includes 19 instances of the token} \textit{wenporo} (``very big''; a combination of \textit{wen}, ``bad(ly)'', and \textit{poro}, ``big''), \hl{whereas in the training data it is transcribed in two different ways:} \textit{wenporo} (11 instances) or \textit{wen poro} (14 instances).
\hl{As a result, both variants can be found in the model's predictions.
In the future, we will investigate methods for automatic detection of discrepancies like this in the data.}

\begin{table}[width=.9\linewidth,cols=4,pos=tp]
\caption{Evaluation results after preprocessing the Viterbi decoder's output and ground truth data by romanizing Japanese characters.
Numbers in brackets indicate the change of CER/WER compared to results in Table~\ref{tab:ft_rslts}.}\label{tab:ft_rslts_all_latin}
\begin{tabular*}{\tblwidth}{@{} LLLL@{} }
\toprule
Fine-tuning data & Additional pretr. steps & CER & WER \\
\midrule
``Wenenekaype'' & 0 & 13.7 (-0.2) & 42.3 \\
Sakhalin Ainu (``Wenenekaype'' + Tuytah) & 0 & 11.6 (-0.2) & 37.3 \\
Sakhalin Ainu + Hokkaido Ainu & 0 & 15.6 (-0.3) & 50.5 (-0.1) \\
``Wenenekaype'' + Hokkaido Ainu & 0 & 16.0 (-0.4) & 51.0 (-0.1) \\
``Wenenekaype'' + Japanese & 0 & 13.4 \textbf{(-2.3)} & 40.9 \textbf{(-0.9)} \\
``Wenenekaype'' + English & 0 & 13.6 (-0.3) & 43.0 \\
``Wenenekaype'' + Hokk. Ainu + Jap. & 0 & 15.8 (-0.2) & 52.0 \\
\hl{``Wenenekaype'' + Hokk. Ainu + Jap. + Eng.} & 0 & 14.8 (-0.4) & 47.4 \\
``Wenenekaype'' & 100k & 10.5 (-0.1) & 33.5 \\
Sakhalin Ainu (``Wenenekaype'' + Tuytah) & 100k & \textbf{9.6} (-0.1) & \textbf{29.8} \\
Sakhalin Ainu + Hokkaido Ainu & 100k & 9.8 (-0.3) & 31.1 \\
``Wenenekaype'' + Hokkaido Ainu & 100k & 9.9 (-0.3) & 30.5 \\
``Wenenekaype'' + Japanese & 100k & 10.1 (-0.7) & 31.9 (-0.3) \\
``Wenenekaype'' + English & 100k & 11.9 (-0.1) & 40.1 \\
``Wenenekaype'' + Hokk. Ainu + Jap. & 100k & 10.3 (-0.6) & 33.5 (-0.2) \\
\hl{``Wenenekaype'' + Hokk. Ainu + Jap. + Eng.} & 100k & 10.9 & 33.9 \\
\bottomrule
\end{tabular*}
\end{table}

\subsubsection{Impact of the Decoding Method}

\cite{Baevski2020wav2vec2A} reported large improvements in WER when decoding their model's output with a textual language model.
As shown in Table~\ref{tab:ft_rslts}, for the model before additional pretraining we also observed significant reduction of WER, but at the cost of higher CER.
For instance, when decoding the model fine-tuned on all Sakhalin Ainu data (``Wenenekaype'' + Tuytah) with the 4-gram model, we achieved a 15\% reduction of WER while at the same time CER increased by nearly 20\%, compared to the Viterbi decoder.
Table~\ref{tab:ex_vit_vs_lm} shows an example of transcriptions generated by using both decoding methods for a single sample from the test data.

\begin{table}[]
\caption{Output of a fine-tuned model for a single test sample, decoded with two different methods.
The KenLM model forces the use of in-vocabulary words (e.g., \textit{reekoh} instead of \textit{reepoh} and \textit{'ekasihi} rather than \textit{'ekasihii}) which often leads to lower Word Error Rates.
On the other hand, Viterbi decoder is better at handling out-of-vocabulary items (e.g., \textit{'ankopuri} is transcribed as \textit{'ankopuuri}, whereas the language model replaced it with a completely different word, \textit{'ankopisi}) which results in lower CER.
}
\label{tab:ex_vit_vs_lm}
\centering
\begin{tabularx}{\textwidth}{rX}
\toprule
Viterbi decoder:  & ne'ohah nay kohnean tani macirih 'oho nay kohne 'anu wa reekoh ne'an cispuurikara 'ankii manuyke 'awwen 'ekasihii 'ireske 'ankamuy henke 'ohta 'ankopuuri 'ahsin manuyke reepoh ne'an henke 'ihunke kii manuyke reekoh  \\
\midrule
KenLM decoder:  & 'oha nay konna tani maciri 'ohonkesehe ne'an cispuurikara 'ankii manuyke 'awwen 'ekasihi 'ireske 'ankamuy henke 'ohta 'ankopisi 'asin manuyke reekoh ne'an henke 'ihunke kii manuyke reekoh \\
\midrule
Ground truth:  & 'ohah naykoh ne 'an tani macirihi 'ohoo naykoh ne 'a nu wa reekoh ne'an cispuurikara 'ankii manuyke 'awwen 'ekasihi 'ireske 'ankamuy henkehe 'ohta 'ankopuri'ahte manuyke reekoh ne'an henke 'ihunke kii manuyke reekoh \\
\bottomrule
\end{tabularx}
\end{table}

Continued pretraining on Ainu data closed the gap in terms of WER between the two decoding methods (see Figure~\ref{fig:effect_of_pretr}) and the best overall results were obtained by decoding without a language model.
This outcome indicates that after having been taught a representation of the target language, a speech representation model is capable of learning an implicit language model from the fine-tuning data which is more powerful than a count-based n-gram model computed from the same data.
A major difference between \cite{Baevski2020wav2vec2A}'s and our setting is that while their language models were trained on a book corpus comprising over 800 million tokens \citep{Librispeech}, Sakhalin Ainu textual data available to us was limited to the transcriptions of the speech data which we also used to fine-tune our model.
This leads to a conclusion that in a language documentation scenario, where large amounts of textual data are typically not available, decoding with an external language model may be of limited use.
However, future work should investigate whether similar trends would also occur when using other types of language models, such as a character-level language model.
Another factor that may be influencing our results is the characteristics of the writing system:
\cite{Baevski2020wav2vec2A} conducted their experiments on English, which exhibits many-to-many correspondences between graphemes and phonemes.
Ainu, on the other hand, is transcribed according to a phonemic orthography,
which reduces the need for explicit information about the correct spelling of individual lexical items.

\subsubsection{Impact of Overlapping Character Vocabulary}

\cite{Conneau2021UnsupervisedCR} found that using a shared phoneme vocabulary in multilingual fine-tuning yields better results than maintaining a separate vocabulary for each language.
While our system is not operating on phonemes, we wanted to verify if an overlap in the character vocabularies representing the target language and other languages used in fine-tuning had an influence on the accuracy of speech transcription.
Specifically, we fine-tuned additional models with (i) Hokkaido Ainu transcriptions converted to upper case letters, (ii) LibriSpeech (English) transcriptions converted to lower case letters, and (iii) Japanese transcriptions converted to lower case alphabet letters.
In all cases we used the model pretrained on Ainu language data.
The output of the models was decoded with a Viterbi decoder.
To minimize the influence of code-switched parts written with Japanese script in Ainu language data, \textit{katakana} characters in the generated transcriptions and ground truth data were transliterated into Latin letters before evaluation.

Results are presented in Table~\ref{tab:ft_rslts_vocab}.
With less than 1h of target language data (i.e., ``Wenenekaype'' only) and additional data from a similar language (Hokkaido Ainu or Japanese), using a shared output vocabulary resulted in lower error rates.
Converting English text to lower case produced mixed results, with improved WER but higher CER.
After increasing the quantity of labeled Sakhalin Ainu data, the opposite outcome was observed: the model trained on Hokkaido Ainu text in upper case yielded better performance.
In contrast to the experiment with shared character vocabulary, in this setting fine-tuning jointly on Sakhalin and Hokkaido Ainu data resulted in a small improvement compared to using Sakhalin Ainu data only.

\begin{table}[width=.9\linewidth,cols=4,pos=tp]
\caption{Effect of using separate or shared character vocabularies for the target language and additional languages represented in fine-tuning.}\label{tab:ft_rslts_vocab}
\begin{tabular*}{\tblwidth}{@{} LLLL@{} }
\toprule
Fine-tuning data & Char. vocabulary size & CER & WER \\
\midrule
``Wenenekaype'' & 76 & 10.5 & 33.5 \\
``Wenenekaype'' (lower case) + Hokk. Ainu (UPPER CASE) & 102 & 10.2 & 31.2 \\
``Wenenekaype'' (lower case) + Hokk. Ainu (lower case) & 86 & 9.9 & 30.5 \\
``Wenenekaype'' (alphabet) + Japanese (\textit{katakana}) & 137 & 10.1 & 31.9 \\
``Wenenekaype'' (alphabet) + Japanese (alphabet) & 119 & 9.7 & 31.0 \\
``Wenenekaype'' (lower case) + English (UPPER CASE) & 102 & 11.9 & 40.1 \\
``Wenenekaype'' (lower case) + English (lower case) & 86 & 12.3 & 39.6 \\
Sakhalin Ainu & 94 & 9.6 & 29.8 \\
Sakh. Ainu (lower case) + Hokk. Ainu (UPPER CASE) & 111 & \textbf{9.4} & \textbf{29.6} \\
Sakh. Ainu (lower case) + Hokk. Ainu (lower case) & 95 & 9.8 & 31.1 \\
\bottomrule
\end{tabular*}
\end{table}

Given these results, we conclude that when there is very little labeled target language data and additional data from a similar speech variety is utilized, it is better to use a shared character vocabulary,
as it seems to facilitate cross-lingual transfer. 
This recommendation should be treated with caution, as similarity between languages is not necessarily matched by similarity in their writing systems (or transliteration methods), and thus the results for other combinations of languages may be different.
On the other hand, as long as the phoneme-grapheme mappings differ in predictable ways, mismatches can be handled by applying rule-based preprocessing to the transcriptions.
With more target language data available,
the model is more often confused by the cross-lingual signal than it benefits from it, and therefore
it would probably be best to just execute monolingual fine-tuning (which also saves compute), but it may be worth experimenting with a separate character vocabulary.
Future research should perform similar analyses for other languages.

\section{Conclusions and Future Work}
\label{sec:conclusions}

We have demonstrated that a strong multilingual speech representation model, such as XLSR-53, can be adapted for a new, low-resource language through multilingual fine-tuning and additional pretraining, resulting in improved downstream performance.
Through experiments with automatic transcription of Sakhalin Ainu, we found that continued pretraining on target language data leads to substantial reduction in error rates.
Furthermore, our results show that in a scenario where labeled target language data is extremely scarce, the model can take advantage of data from a related speech variety (or -- to a lesser extent -- an unrelated language with similar phonological traits) added during fine-tuning,
if that additional language was seen during pretraining.

Our findings confirm the hypothesis that language similarity should be taken into consideration and can be leveraged in the process of multilingual fine-tuning.
They also indicate that self-supervised pretraining of a language representation model is not only effective in adapting it for a particular language, but the representations learned during that process can also serve as a bridge for transfer to similar languages in the form of cross-lingual supervision.
We expect this observation to also be true for cross-domain supervision within the same language -- future work should investigate this assumption.

\hl{As a next step in our research, we are planning to increase the amount of labeled in-domain data by digitizing and aligning speech data and transcriptions from} \cite{Murasaki1976} and \cite{Murasaki2013_sentences1,Murasaki2016_sentences2}.
\hl{Apart from that, we will test other types of language models for decoding (specifically, neural and character-level language models).}
\hl{Furthermore, we will examine potential methods for reducing negative cross-lingual signal while retaining as much as possible of the benefits, such as fine-tuning with language embeddings and ensemble models.
We also plan to explore data augmentation techniques.}

\section*{Acknowledgements}
This work was supported by JSPS KAKENHI Grant Number JP22K17952.

\bibliographystyle{cas-model2-names}

\bibliography{bibliography}

\end{document}